\documentclass[letterpaper, 10 pt, conference]{ieeeconf}  

\usepackage{grffile}
\usepackage{amsmath}
\usepackage{graphicx} 
\usepackage{graphics} 
\usepackage{cite}
\usepackage{stfloats}
\usepackage{float} 
\usepackage{bm}

\IEEEoverridecommandlockouts                              

\title{\LARGE \bf
Bioinspired Kirigami Capsule Robot for Minimally Invasive Gastrointestinal Biopsy
}

\author{Ruizhou Zhao$^{1,\dagger}$, Yichen Chu$^{2,\dagger}$, Shuwei Zhao$^{1}$, Wenchao Yue$^{1}$,\\
Raymond Shing-Yan Tang$^{3}$, and Hongliang Ren$^{1,*}$
\thanks{The work was supported by the Guangdong Basic and Applied Basic Research Foundation (GBABF) under Grant 2021B1515120035, Hong Kong Research Grants Council (RGC)  Collaborative Research Fund (C4026-21G), Research Grants Council (RGC) - Research Impact Fund (RIF R4020-22), and General Research Fund (GRF 14203323).
(\textit{Corresponding Author: Hongliang Ren})
}
\thanks{$^{1}$R. Zhao, S. Zhao, W. Yue, and H. Ren are with the Department of Electronic Engineering, The Chinese University of Hong Kong, Sha Tin, Hong Kong, China, and CUHK Shenzhen Research Institute, CUHK-SZRI, Shenzhen, China.
       {\tt\small 1155225302@link.cuhk.edu.hk; hlren@ieee.org;}}%
\thanks{$^{2}$Y. Chu is with the Department of Mechanical Engineering and Automation, Northeastern University, Shenyang 110819, China.
        {\tt\small 2310088@stu.neu.edu.cn}}%
\thanks{$^{3}$Raymond Shing-Yan Tang is with the Department of Medicine and Therapeutics and Institute of Digestive Disease, The Chinese University of Hong Kong, Hong Kong.}%
\thanks{$^{\dagger}$These two authors contributed equally to this work.}%
}

\begin{document}

\maketitle
\thispagestyle{empty}
\pagestyle{empty}


\begin{abstract}
Wireless capsule endoscopy (WCE) has transformed gastrointestinal (GI) diagnostics by enabling noninvasive visualization of the digestive tract, yet its diagnostic yield remains constrained by the absence of biopsy capability, as histological analysis is still the gold standard for confirming disease. Conventional biopsy using forceps, needles, or rotating blades is invasive, limited in reach, and carries risks of perforation or mucosal trauma, while fluid- or microbiota-sampling capsules cannot provide structured tissue for pathology, leaving a critical gap in swallowable biopsy solutions. Here we present the Kiri-Capsule, a kirigami-inspired capsule robot that integrates deployable PI-film flaps actuated by a compact dual-cam mechanism to achieve minimally invasive and repeatable tissue collection. The kirigami surface remains flat during locomotion but transforms into sharp protrusions upon cam-driven stretching, enabling controlled penetration followed by rotary scraping, with specimens retained in internal fan-shaped cavities. Bench tests confirmed that PI films exhibit a Young’s modulus of approximately 20 MPa and stable deployment angles (about $\bm{34^{\circ}}$ at 15\% strain), while ex vivo porcine studies demonstrated shallow penetration depths (median $\sim$0.61 mm, range 0.46–0.66 mm) and biopsy yields comparable to standard forceps (mean $\sim$10.9 mg for stomach and $\sim$18.9 mg for intestine), with forces within safe ranges reported for GI biopsy. These findings demonstrate that the Kiri-Capsule bridges passive imaging and functional biopsy, providing a swallowable, depth-controlled, and histology-ready solution that advances capsule-based diagnostics toward safe and effective clinical application.
\end{abstract}


\section{Introduction}
Wireless capsule endoscopy (WCE) has transformed gastrointestinal (GI) diagnostics by enabling noninvasive visualization, yet its value is limited by the lack of tissue acquisition \cite{cao2024robotic, mehedi2023intelligent, saurin2016challenges, iddan2000wireless}. Histological biopsy remains the gold standard for confirming early cancer, inflammatory bowel disease, and Helicobacter pylori–associated gastritis \cite{mercado2022gut, mansell2003biopsy}; without biopsy, capsule-based findings are not definitive. Conventional endoscopic biopsy (forceps or needle puncture) can obtain adequate samples but is invasive, uncomfortable, and poorly suited to deep small-intestinal regions \cite{min2020robotics, kong2012robotic, achkar1986comparison}. Recent capsule-robotics strategies—needle modules, rotating blades, scraping tools, and multibarrel designs—demonstrate feasibility but face risks of perforation or mucosal trauma, limited control of penetration depth, variable yields, and added structural complexity that compromises swallowability \cite{abdigazy2024end, son2020magnetically, hoang2019untethered, chen2014design}. Fluid- or microbiota-collection capsules have been reported, yet they lack the structured tissue required for pathology \cite{del2024soft, park2024multiple, rezaei2019ingestible}. These limitations motivate biopsy-capable capsule robots that combine ingestibility with safe and repeatable collection of GI tissue \cite{cao2024robotic, min2020robotics}.

\begin{figure}[h]
  \centering
  \includegraphics[width=0.51\textwidth]{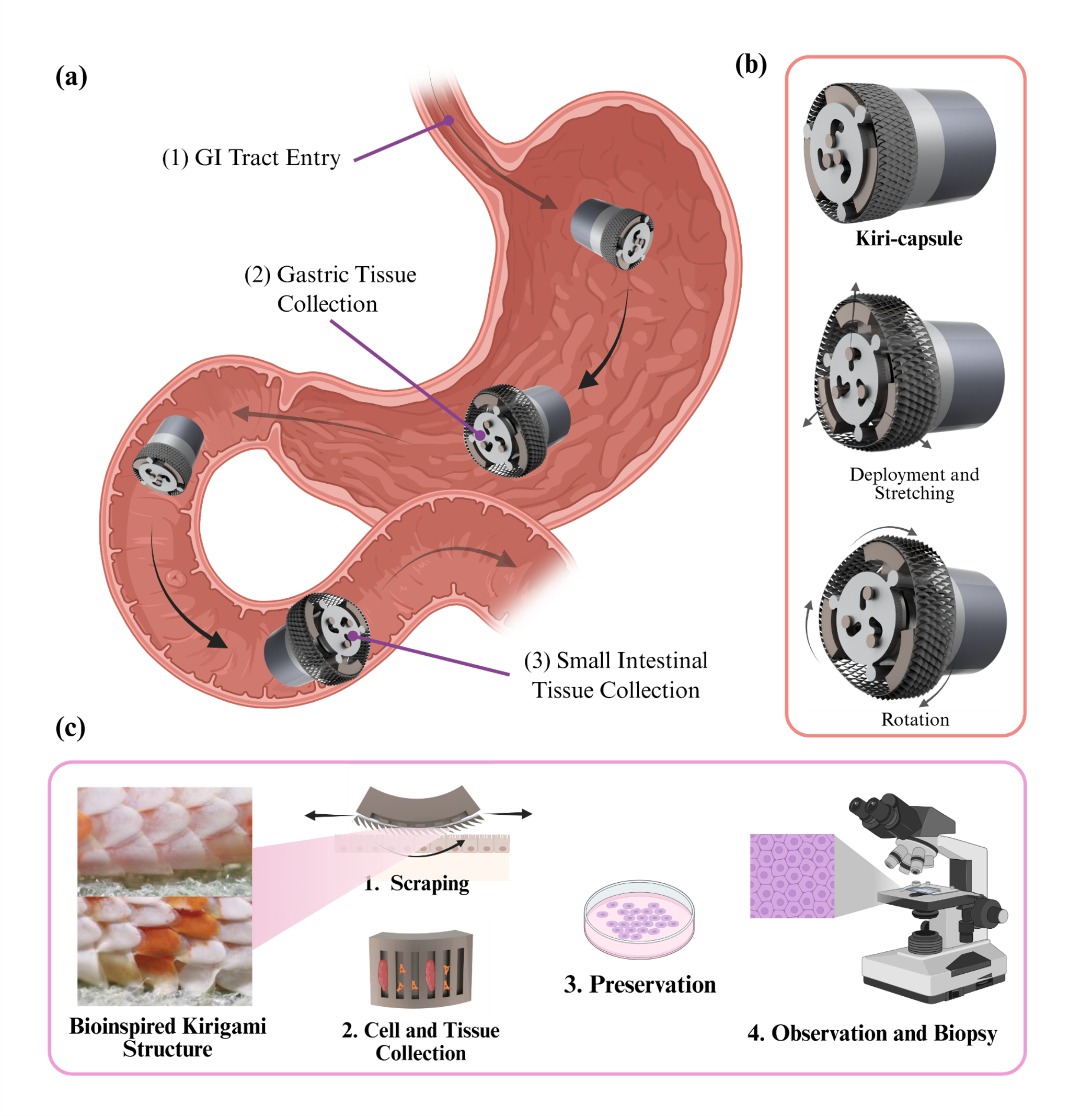}
  \caption{Overview of the proposed kiri-capsule system. (a) The capsule is ingested and performs biopsy in the stomach and small intestine. (b) Working principle: the kirigami skin deploys via dual-cam expansion, followed by rotational scraping for tissue detachment. (c) Bioinspired kirigami scales mimic snake skin, enabling tissue collection on the flap surface and storage within fan-shaped internal cavities for subsequent histological analysis.}
    \label{fig1}
\end{figure}

Kirigami-inspired structures, formed by patterned cuts in thin films to generate stretch-induced three-dimensional protrusions, offer a promising pathway \cite{babu2024tailoring, brooks2022kirigami, yang2021grasping}. Unlike rigid puncture tools, kirigami flaps can be deployed on demand to create depth-bounded protrusions that engage tissue more safely; coupled with rotary actuation, they scrape superficial mucosa while reducing perforation risk \cite{babaee2021kirigami}. The same flaps can capture and retain tissue on their surfaces or within cavities, enabling reliable transfer for histological analysis. Such properties position kirigami design as a strong candidate for addressing the long-standing challenge of safe and effective GI biopsy in ingestible capsule robots.

To address these limitations, we propose the Kiri-Capsule, a kirigami-inspired biopsy capsule robot that integrates soft actuation and bioinspired deployable structures for controlled GI tissue collection (Fig.~\ref{fig1}a--c). The kirigami surface patterns, fabricated from biocompatible PI film, remain flat during locomotion but transform into protruding sharp flaps under dual-cam expansion. The design of these flaps is biomimetically inspired by the overlapping motion of snake scales. Once actuated, the capsule’s front-end actuator rotates to scrape and detach mucosal tissues, which are then stored within internal compartments for subsequent histological analysis. This architecture enables controlled tissue acquisition without high-penetration forces, reducing perforation risk while preserving ingestible dimensions.

The contributions of this work are summarized as follows:

\begin{itemize}
    \item We design a kirigami-inspired biopsy capsule robot that introduces controllable surface transformations to achieve safe and minimally invasive GI tissue sampling.
    \item We develop a dual-cam expansion mechanism that drives kirigami protrusions in a compact ingestible form factor, enabling repeatable and reversible flap deployment.
    \item We demonstrate rotary scraping-based collection with encapsulation inside the capsule body, ensuring reliable specimen retrieval and reduced tissue trauma.
    \item Through benchtop and ex vivo validation, we show that the Kiri-Capsule achieves penetration depths and biopsy yields comparable to clinical benchmarks, while maintaining forces within safe ranges for gastric and intestinal tissues.
\end{itemize}

\section{Design, Fabrication, and Assembly of the Kiri-Capsule}

\begin{figure*}[t]
  \centering
  \includegraphics[width=6.3in]{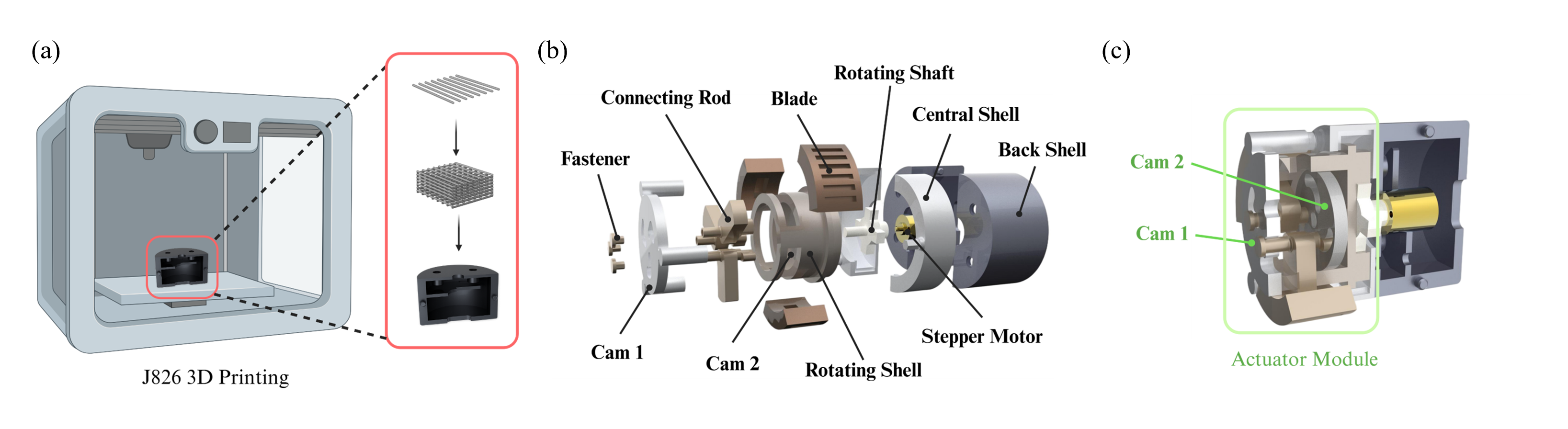}
  \caption{Design and structural assembly of the kiri-capsule. (a) Fabrication of capsule body parts using a J826 stereolithography printer. (b) Exploded view showing major components, including fastener, cam~1, cam~2, rotating shell, stepper motor, back shell, central shell, rotating shaft, blade, and connecting rod. (c) Cross-sectional view of the assembled capsule with the anterior actuator driven by dual cams and a miniature stepper motor, while the posterior shells protect the motor and electronics.}
    \label{fig2}
\end{figure*}

\subsection{Fabrication and Structural Assembly}

The fabrication of the kiri-capsule body and mechanical components was carried out using a J826 photopolymer 3D printer (Stratasys, USA), as shown in Fig.~\ref{fig2}a. This high-resolution stereolithography-based system provides accurate reproduction of small-scale geometries and smooth surfaces, which are critical for mechanical robustness and biocompatibility. The additive manufacturing process enabled rapid prototyping of the outer shell, cam components, and housing parts, ensuring dimensional consistency for subsequent assembly.

An exploded view of the complete system is presented in Fig.~\ref{fig2}b. The capsule consists of a fastener, dual cams (cam~1 and cam~2), a rotating shell, a miniature stepper motor, the back and central shells, a rotating shaft, a blade, and a connecting rod. Each element is designed to support compact integration while maintaining functional reliability. In particular, the connecting rod is designed in a non-stacked configuration, which constrains the flaps during operation and prevents lateral wobbling, thereby ensuring smooth rotational motion.

A cross-sectional view of the fully assembled capsule is illustrated in Fig.~\ref{fig2}c. The anterior section houses the biopsy actuator, which relies on a dual-cam expansion mechanism. When the cams rotate, the kirigami flaps expand outward in a controlled manner. This motion is driven by a miniature four-phase stepper motor, which transfers torque through a cross-shaped shaft to the front-end actuator. The actuator subsequently performs rotational scraping of mucosal tissue. To protect the motor and electronic components from exposure to the gastrointestinal environment, the central and posterior shells enclose the assembly, providing both structural integrity and environmental isolation. Finally, the entire capsule is hermetically packaged to ensure safe ingestion and robust operation within the GI tract.

\subsection{Working Principle and Cam-Driven Kinematics}

\begin{figure}[t]
  \centering
  \includegraphics[width=0.5\textwidth]{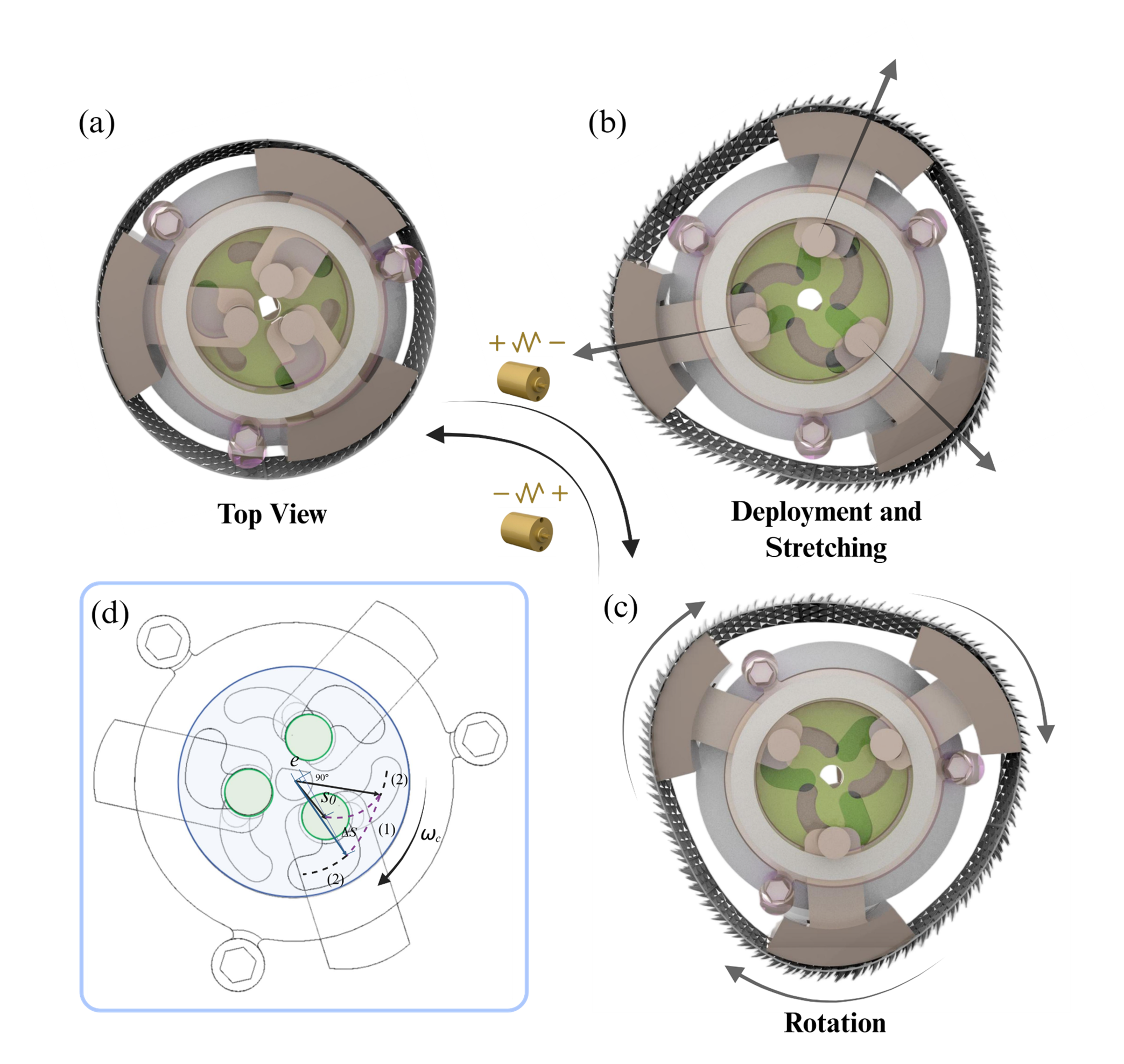}
  \caption{Working principle and cam kinematics of the kiri-capsule. (a) Initial top view. (b) Deployment and stretching of the kirigami flaps via cam-driven plate translation. (c) Rotational scraping as the plates follow the cam groove. (d) Kinematic model and geometric parameters of the profiled cam–expansion plate mechanism.}
  \label{fig3}
\end{figure}

Figure~\ref{fig3} summarizes the operation of the kiri-capsule actuator, which is driven by a pair of profiled (curvilinear) cams. The actuation sequence consists of two phases. During the \emph{deployment and stretching} phase (Fig.~\ref{fig3}a,b), the lower arc-shaped cam rotates while the upper follower is constrained to translate along a straight guide. This motion drives the expansion plates outward, stretching the kirigami skin and transforming the patterned PI film into protruding sharp flaps along the predefined cut paths. In the subsequent \emph{rotational scraping} phase (Fig.~\ref{fig3}c), continued cam rotation guides the follower into the curved groove of the upper cam, inducing a clockwise rotation of the expansion plates. This action produces effective scraping while the kirigami flaps remain engaged with the gastrointestinal wall. Torque is supplied by a miniature four-phase stepper motor through a cross-shaped shaft, enabling precise and reversible control. Positive motor polarity yields deployment followed by scraping, whereas reversing the polarity drives flap retraction and resets the mechanism.

The actuator is modeled as a profiled cam with an offset translating roller follower (Fig.~\ref{fig3}d). The cam rotates with angular velocity \(\omega_c\), while the follower is constrained to vertical translation. Let the lateral offset and initial vertical distance be \(e\) and \(s_0\), respectively, and define the cam angle \(\phi=\omega_c t\). The follower displacement law \(s(\phi)\) prescribes its motion, and the corresponding pressure angle is

\vspace{-1mm}
\begin{equation}
    \mu(\phi) = \arctan\!\left(\frac{e}{s_0+s(\phi)}\right).
\end{equation}
\vspace{-1mm}
The follower kinematics then follow as
\vspace{-1mm}
\begin{align}
    y(\phi) &= s_0+s(\phi), \\
    \dot{y}(\phi) &= \frac{ds}{d\phi}\,\omega_c, \\
    \ddot{y}(\phi) &= \frac{d^2s}{d\phi^2}\,\omega_c^2.
\end{align}

By appropriately designing \(s(\phi)\) (e.g., cycloidal, modified sine, or polynomial with jerk limits) and selecting a constant or trapezoidal \(\omega_c(t)\), the deployment and scraping motions can be shaped to maintain bounded pressure angles and accelerations, thereby achieving smooth flap expansion, stable tissue engagement, and reliable retraction.

\subsection{Kirigami Design and Fabrication}

\begin{figure}[h]
  \centering
  \includegraphics[width=0.48\textwidth]{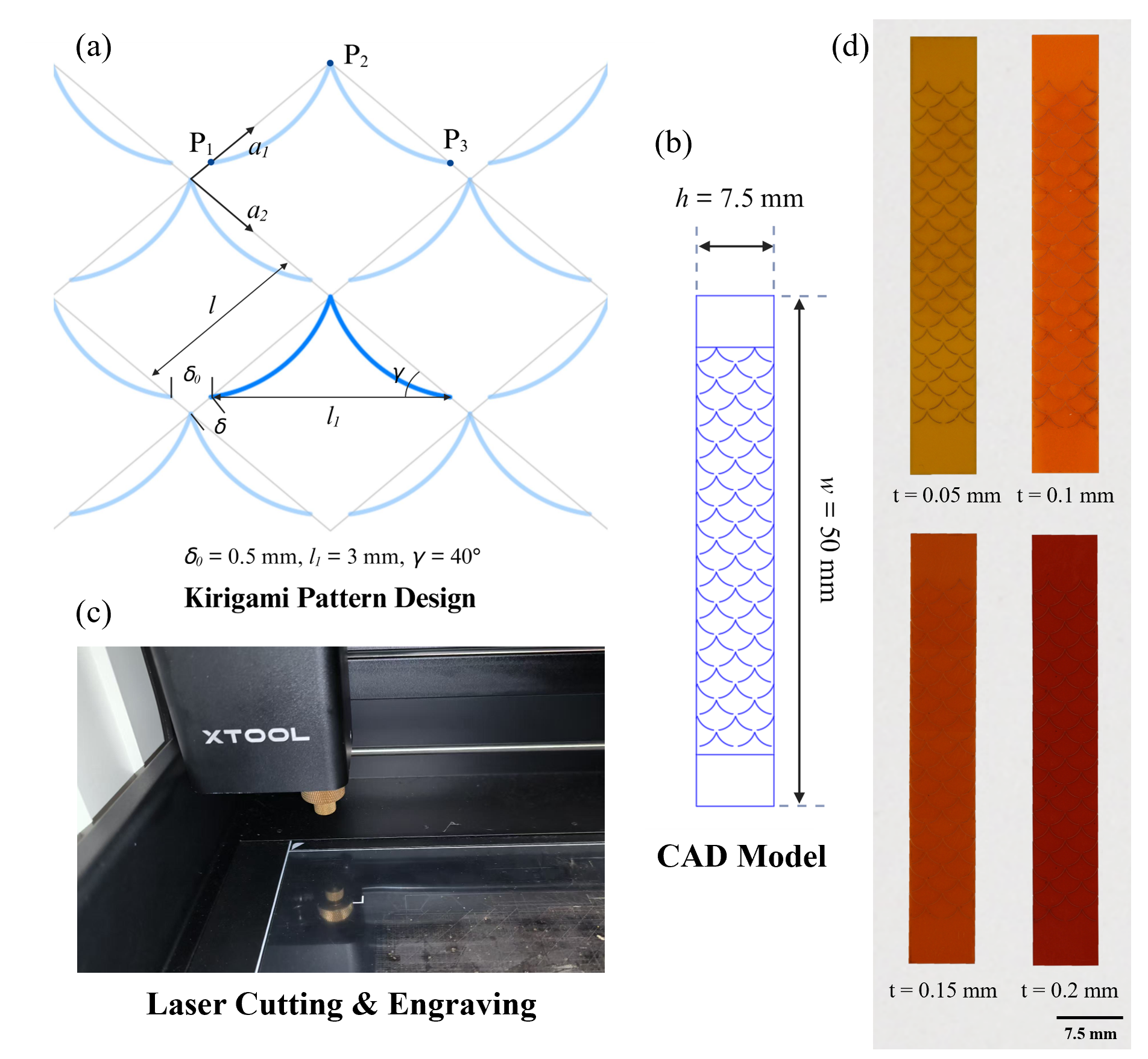}
  \caption{Design and fabrication of the kirigami biopsy surface. (a) Kirigami pattern with parameters $\delta_0 = 0.5$~mm, $l_1 = 3$~mm, and $\gamma = 40^{\circ}$. (b) CAD model with strip dimensions $h = 7.5$~mm and $w = 50$~mm. (c) Laser cutting and engraving process using PI film. (d) Fabricated kirigami sheets with four different PI thicknesses ($t = 0.05, 0.1, 0.15, 0.2$~mm).}
  \label{fig4}
\end{figure}

Figure~\ref{fig4} illustrates the kirigami-inspired biopsy skin integrated into the kiri-capsule. The cuts are arranged on a triangular lattice, with the unit cell defined by the primitive vectors
\begin{equation}
  \mathbf{a}_1=\begin{bmatrix} l\cos\gamma \\[2pt] l\sin\gamma \end{bmatrix}, \qquad
  \mathbf{a}_2=\begin{bmatrix} l\cos\gamma \\[2pt] -\,l\sin\gamma \end{bmatrix}.
\end{equation}

Within each unit cell, the representative cut is specified by the endpoints

\begin{equation}
  P_1=\tfrac{\delta}{l}\,\mathbf{a}_1, \qquad
  P_3=\mathbf{a}_1+\Bigl(1-\tfrac{\delta}{l}\Bigr)\mathbf{a}_2,
\end{equation}

where $P_1$ marks the initial notch location and $P_3$ the terminal point of the slit. The triangular protrusion units are defined by three key parameters: notch length $l_1$, cut spacing $\delta_0$, and opening angle $\gamma$. These geometric factors jointly determine the in-plane stretchability of the sheet as well as the extent of out-of-plane protrusion needed for effective mucosal engagement. For the present design, we selected $\delta_0=0.5$~mm, $l_1=3$~mm, and $\gamma=40^{\circ}$ to achieve a balance between penetration sharpness and structural robustness. The overall strip dimensions were specified as $h=7.5$~mm and $w=50$~mm (Fig.~\ref{fig4}b), ensuring full compatibility with the capsule actuator geometry.

The kirigami sheets were fabricated using laser cutting (XTOOL) on polyimide (PI) films, chosen for their favorable mechanical properties—sufficient hardness to assist puncture and moderate extensibility to withstand repeated actuation, as illustrated in Fig.~\ref{fig4}c. Four thicknesses ($t=0.05, 0.1, 0.15, 0.2$~mm; Fig.~\ref{fig4}d) were prepared to investigate the influence of film thickness on protrusion amplitude, stiffness, and tissue interaction. This systematic variation provides a basis for optimizing the biopsy performance of the kiri-capsule.

\subsection{Prototype Implementation}

\begin{figure}[h]
  \centering
  \includegraphics[width=0.43\textwidth]{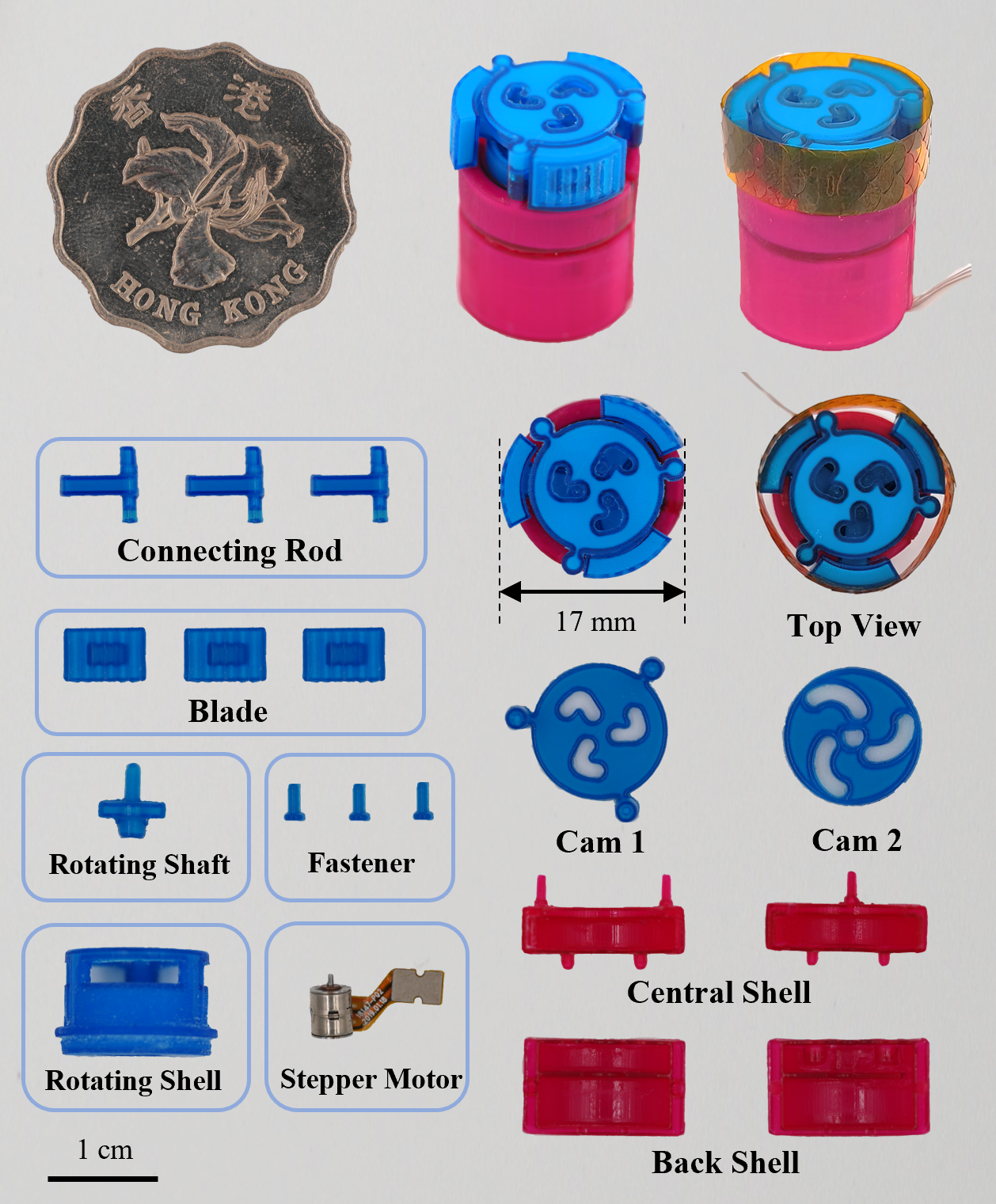}
  \caption{Fabricated prototype of the kiri-capsule. All manufactured parts are displayed, including fastener, cam~1, cam~2, rotating shell, stepper motor, back and central shells, rotating shaft, blade, and connecting rod. The fully assembled capsule with kirigami skin and testing motor is also shown. The capsule dimensions are 17~mm in diameter and 22~mm in length, with a scale bar of 1~cm for reference.}
  \label{fig5}
\end{figure}

Figure~\ref{fig5} presents the fabricated prototype of the kiri-capsule together with all machined components. For scale reference, a Hong Kong one-dollar coin is included in the images. The complete set of manufactured parts is shown, comprising the fastener, cam~1, cam~2, rotating shell, stepper motor, back shell, central shell, rotating shaft, blade, and connecting rod. These components were subsequently assembled into the final capsule structure, which integrates the kirigami biopsy skin and a testing motor for actuation evaluation. The fully assembled capsule achieves an outer diameter of approximately 17~mm and a total length of 22~mm, dimensions that are consistent with clinically relevant ingestible devices. A 1~cm scale bar is provided in Fig.~\ref{fig5} for reference. The prototype highlights the compact integration of mechanical components and the kirigami biopsy module within a swallowable capsule form factor.

In summary, this section has described the structural design, fabrication, and assembly of the kiri-capsule, encompassing the dual-cam actuation mechanism, kirigami surface integration, and compact prototype realization. The incorporation of laser-fabricated PI kirigami sheets into the dual-cam expansion system enables controllable deployment and rotational scraping within a capsule-scale platform. The final prototype, with a diameter of 17~mm and a length of 22~mm, demonstrates the feasibility of achieving a swallowable form factor while preserving functional biopsy capability. The subsequent section will assess the performance of the kiri-capsule through benchtop and ex vivo experiments.

\section{Experimental Testing and Validation}

This section presents the experimental evaluation of the kiri-capsule to validate its mechanical performance, actuation behavior, and biopsy capability. A series of benchtop and ex vivo experiments were conducted to characterize the kirigami sheets, the actuator system, and the tissue interaction outcomes. Specifically, we first investigated the tensile properties of kirigami membranes with different thicknesses and quantified their opening angles under strain. We then calibrated the stepper motor control and examined the penetration depth of the kirigami flaps into soft tissue. Finally, force measurements were performed to assess the push-open and penetration loads during interaction with gastric and small intestinal specimens. Together, these results provide a comprehensive validation of the kirigami design and confirm its feasibility as a minimally invasive biopsy capsule.

\subsection{Tensile Characterization of Kirigami Sheets}

\begin{figure}[h]
  \centering
  \includegraphics[width=0.49\textwidth]{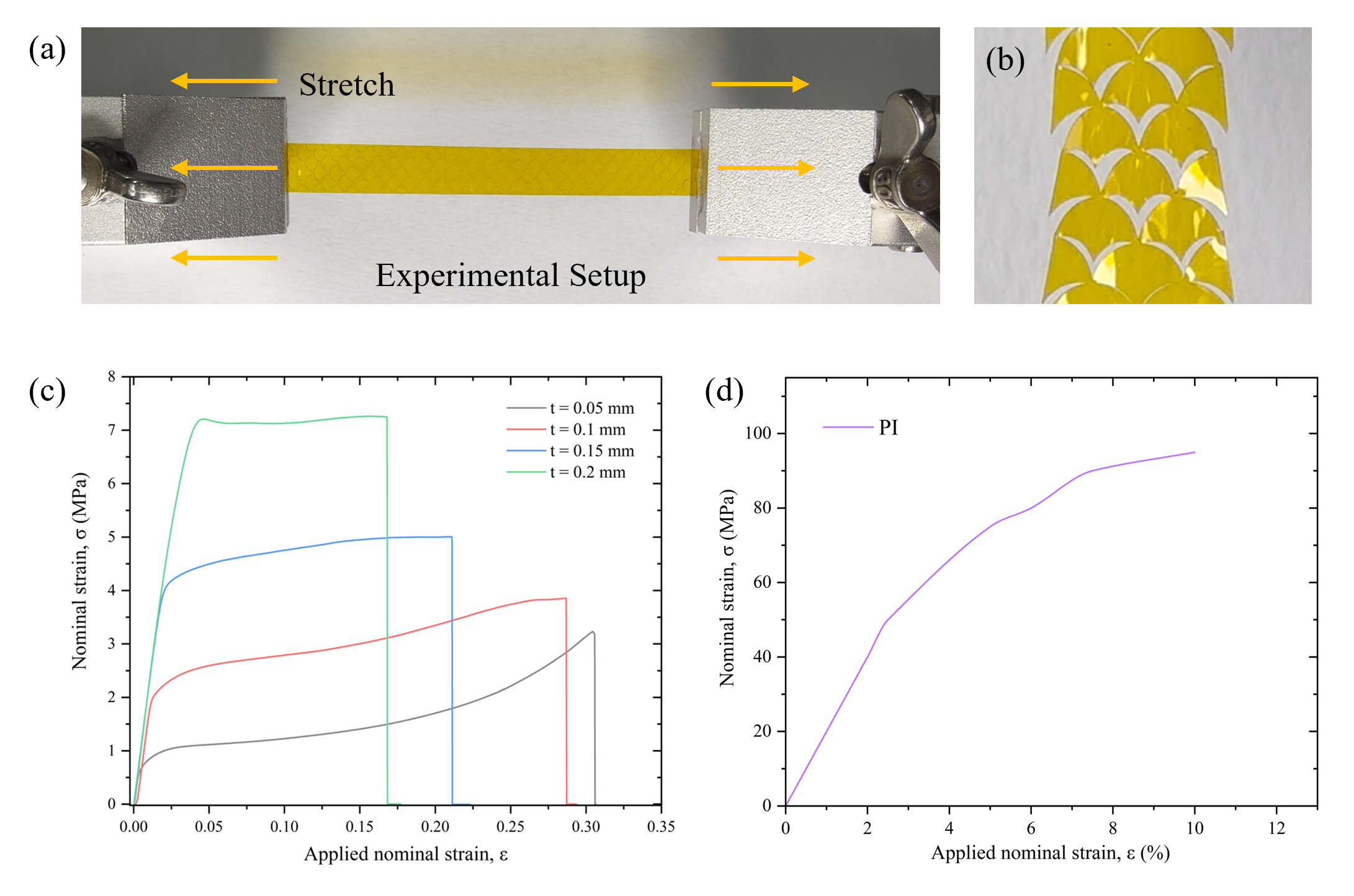}
  \caption{Tensile characterization of kirigami sheets. (a) Experimental setup using a universal tensile testing machine. (b) Stretching of the kirigami PI film showing out-of-plane protrusions. (c) Stress–strain curves of films with different thicknesses ($t = 0.05, 0.1, 0.15, 0.2$~mm). (d) Stress–strain response of pristine PI film, with a measured Young’s modulus of approximately 20~MPa.}
    \label{fig6}
\end{figure}

To evaluate the mechanical properties of the kirigami surface, uniaxial tensile tests were conducted on polyimide (PI) films with different thicknesses. Figure~\ref{fig6}a shows the experimental setup, where a universal tensile testing machine (Xin Sansi, Shanghai, China) was used to stretch the specimens. The kirigami films were clamped at both ends and subjected to controlled displacement loading. As illustrated in Fig.~\ref{fig6}b, the kirigami cuts gradually opened during stretching, producing out-of-plane deformations and exposing sharp protrusions on the surface. This mechanism has been widely observed in kirigami-based metamaterials, where the transition from in-plane linear deformation to out-of-plane shape morphing enables anisotropic mechanical responses.

The stress–strain results are presented in Fig.~\ref{fig6}c for films of four thicknesses ($t = 0.05, 0.1, 0.15, 0.2$~mm). As the thickness increased, the films exhibited higher stiffness but required significantly larger forces to overcome the initial in-plane deformation before entering the softening regime associated with kirigami morphing. For the thicker samples ($t = 0.15$ and $0.2$~mm), the required force was excessively high, which is undesirable for capsule-scale actuation. In contrast, the thinner sheets ($t = 0.05$ and $0.1$~mm) demonstrated smoother transitions to the out-of-plane deformation regime, making them more suitable for integration. The measured Young’s modulus of the PI base film, shown in Fig.~\ref{fig6}d, was approximately 20~MPa, consistent with literature values for polyimide.

From these results, we conclude that film thickness strongly influences the actuation performance of kirigami patterns. Among the tested samples, the 0.05~mm thick kirigami sheet was selected for subsequent experiments, as it provided the optimal balance between mechanical compliance, protrusion effectiveness, and ease of actuation within the capsule.

\subsection{Angular Characterization of Kirigami Deployment}

\begin{figure}[h]
  \centering
  \includegraphics[width=0.49\textwidth]{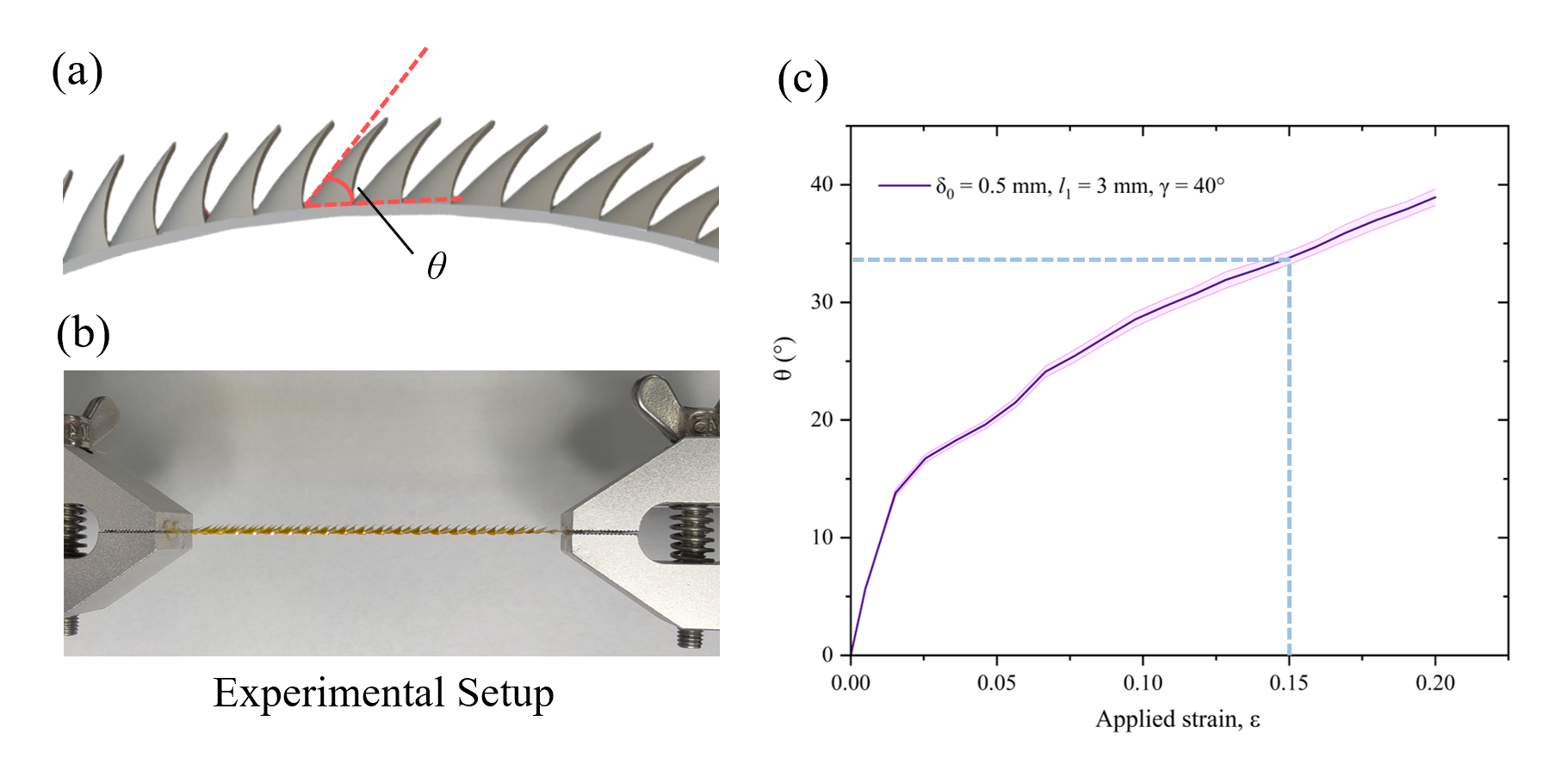}
  \caption{Angular characterization of kirigami deployment. (a) Definition of opening angle $\theta$. (b) Experimental setup for tensile-induced flap deployment. (c) Measured relationship between applied strain and flap angle. At $\varepsilon = 0.15$, the angle was approximately $34^{\circ}$, while at $\varepsilon = 0.20$ the flaps reached $38^{\circ}$. Shaded regions indicate standard error across three experimental trials.}
  \label{fig7}
\end{figure}

To evaluate the deployment behavior of the kirigami flaps, a series of tensile tests was performed to characterize the relationship between applied strain and the opening angle $\theta$. Figure~\ref{fig7}a defines $\theta$ as the angle formed between the tangent of the base film and the tip of the protruding flap. The experimental setup is shown in Fig.~\ref{fig7}b, where kirigami sheets were clamped at both ends and subjected to uniaxial stretching using a universal tensile testing machine. The flap angles were recorded by direct optical observation and subsequently calibrated from video data.  

The results are summarized in Fig.~\ref{fig7}c. For each strain level, 20 independent measurements of $\theta$ were collected, and three experimental repetitions were performed. The solid curve indicates the mean values, while the shaded region denotes the standard error. The flap angle increased nearly linearly with strain before gradually approaching saturation at higher levels of deformation. At a nominal strain of $\varepsilon = 0.15$, corresponding to the actuation range of the capsule system, the kirigami flaps achieved an average opening angle of approximately $34^{\circ}$. At $\varepsilon = 0.20$, the angle increased further to about $38^{\circ}$. These findings confirm that the kirigami surface provides sufficient protrusion within the capsule’s actuation range, ensuring reliable engagement with gastrointestinal tissue during deployment.

\subsection{Electrical Control Principle and Motor Calibration}

\begin{figure}[h]
  \centering
  \includegraphics[width=0.49\textwidth]{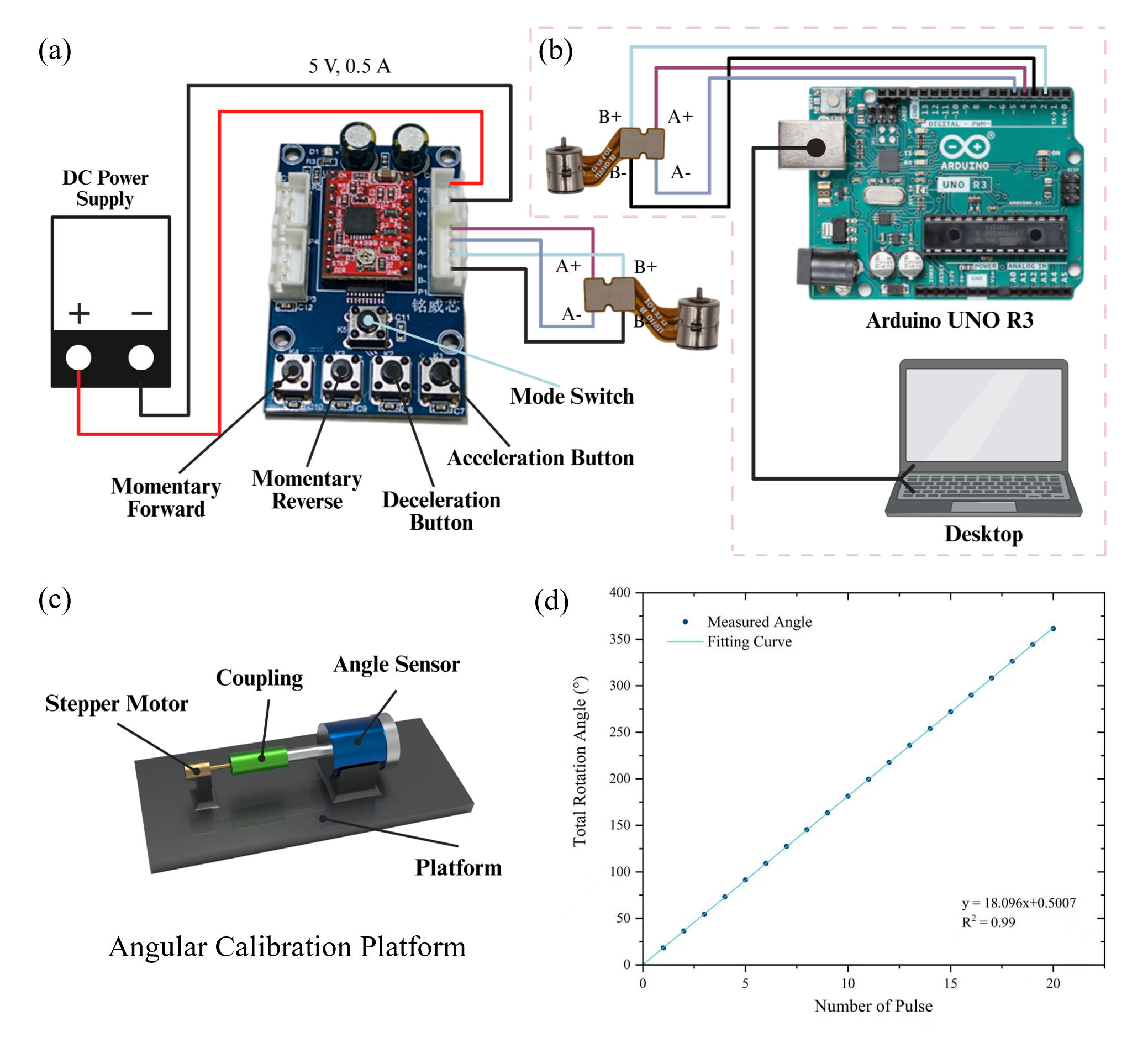}
  \caption{Electrical control and motor calibration. (a) PCB-based motor driver (MINGWEIXIN D194061) with manual control buttons for acceleration, deceleration, forward/reverse jogging, and mode selection. (b) Arduino Uno R3 platform for programmable pulse-based control. (c) Angular calibration platform comprising stepper motor, coupling, and angle sensor. (d) Relationship between motor rotation angle and input pulses, showing linear fitting with $R^2=0.99$.}
  \label{fig8}
\end{figure}

Two distinct methodologies were employed to control the stepper motor used in the kiri-capsule, as illustrated in Fig.~\ref{fig8}a and Fig.~\ref{fig8}b. The first approach utilizes a dedicated printed circuit board (PCB, model MINGWEIXIN D194061) integrating an A4988 motor driver. The PCB is powered by a 5~V, 0.5~A direct current (DC) supply and connected to the motor through four output wires. Five buttons on the PCB provide manual control functions: K1 for acceleration, K2 for deceleration, K3 for momentary reverse jogging, K4 for momentary forward jogging, and K5 for mode selection. By manually actuating these buttons, precise control of motor rotation can be achieved.

The second approach involves an Arduino Uno R3 microcontroller connected to the four motor terminals via its digital I/O pins. Custom software is uploaded to the Arduino to generate pulse-width modulation (PWM) signals, thereby enabling automated and programmable control of the motor. This setup provides flexibility in defining motion sequences and allows synchronization with other capsule subsystems.

To establish the relationship between the motor input and actual angular displacement, an angular calibration platform was constructed (Fig.~\ref{fig8}c). The motor was coupled to an angle sensor mounted on a rigid platform, enabling precise measurement of output rotation. Figure~\ref{fig8}d shows the calibration results, indicating a linear correlation between the number of driving pulses and the total rotation angle. A fitting line with a slope of approximately $18.1^{\circ}$ per pulse and a coefficient of determination $R^2=0.99$ confirms the accuracy and repeatability of the control system. This calibration ensures reliable translation between electrical commands and mechanical actuation in subsequent experiments.

\subsection{Penetration Depth Characterization}

\begin{figure}[h]
  \centering
  \includegraphics[width=0.49\textwidth]{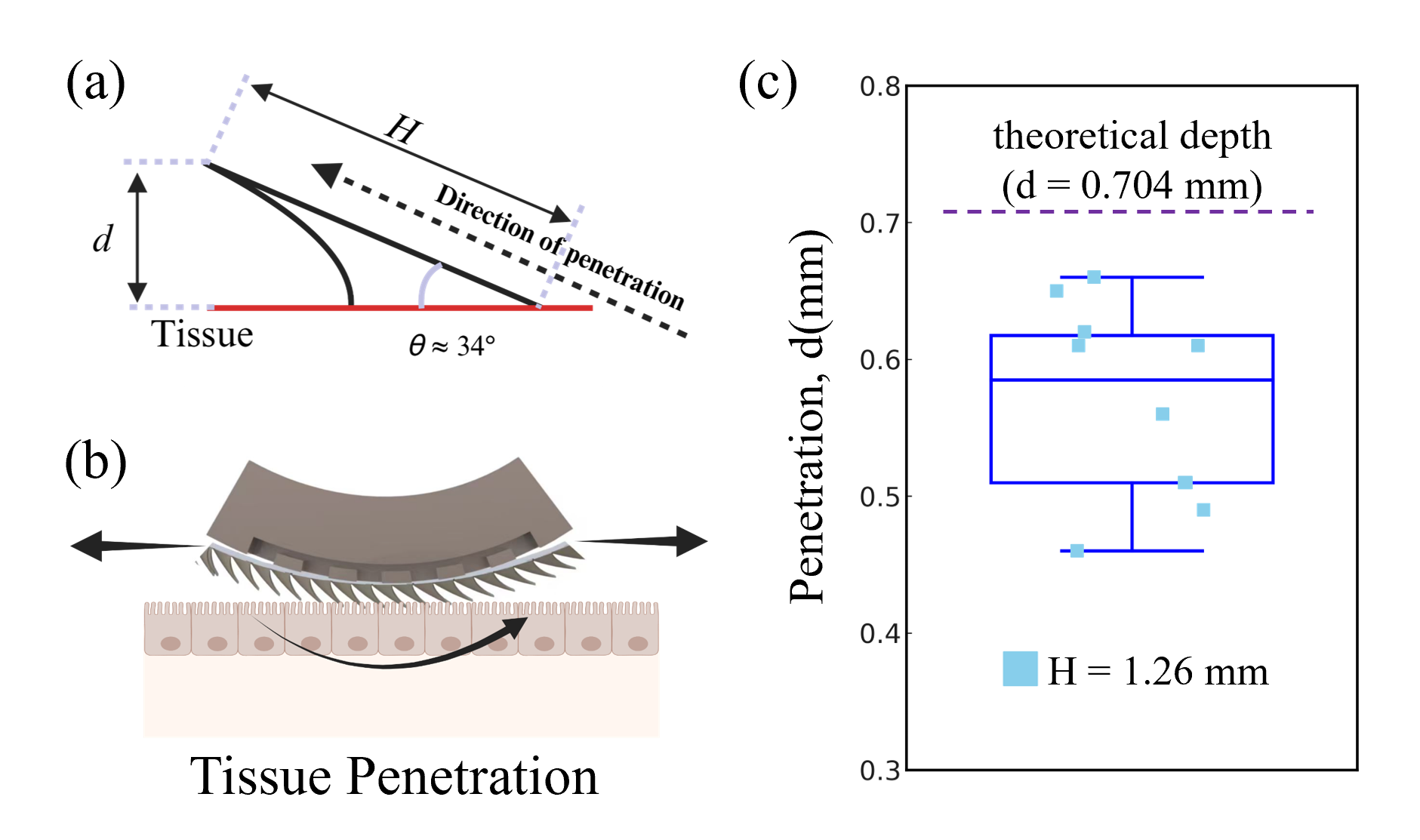}
  \caption{Penetration depth characterization of the kirigami flaps. (a) Geometric definition of spike length $H$ and penetration depth $d$. (b) Schematic of tissue penetration during stretching and scraping. (c) Measured penetration depths in ex vivo porcine tissue ($n=10$), with median depth $\approx 0.61$~mm and theoretical depth $d=0.704$~mm.}
  \label{fig9}
\end{figure}

To evaluate the penetration performance of the kirigami flaps, we measured the depth of tissue insertion under controlled deployment and scraping. Figure~\ref{fig9}a illustrates the geometric definition of the penetration process, where the spike length $H$ is determined by the kirigami geometry ($H = \tfrac{1}{2}\tan(l-\delta)$), and the theoretical penetration depth $d$ can be estimated from the opening angle ($\theta \approx 34^{\circ}$). Figure~\ref{fig9}b shows the conceptual operation principle, in which the kirigami flaps, once deployed, engage the tissue surface and penetrate during the combined stretching and rotational scraping motion.  

The experimental penetration depth results are summarized in Fig.~\ref{fig9}c using a box-and-whisker plot. Across ten ex vivo porcine tissue measurements, the median depth was approximately 0.61~mm (interquartile range 0.51--0.65~mm), with values spanning 0.46--0.66~mm and a mean of 0.57~mm. These values are in close agreement with the theoretical prediction of 0.704~mm derived from the geometric model. Data are presented as the median $\pm$ interquartile range (boxes), with whiskers indicating the smallest and largest values, and cross markers denoting the mean. The consistency between measured and theoretical results confirms that the kirigami flaps can reproducibly achieve shallow, controlled penetration into soft tissue—sufficient for surface biopsy collection while minimizing the risk of excessive injury—and validates the underlying geometric design model.

\subsection{Force and Penetration Characterization}

\begin{figure}[h]
  \centering
  \includegraphics[width=0.49\textwidth]{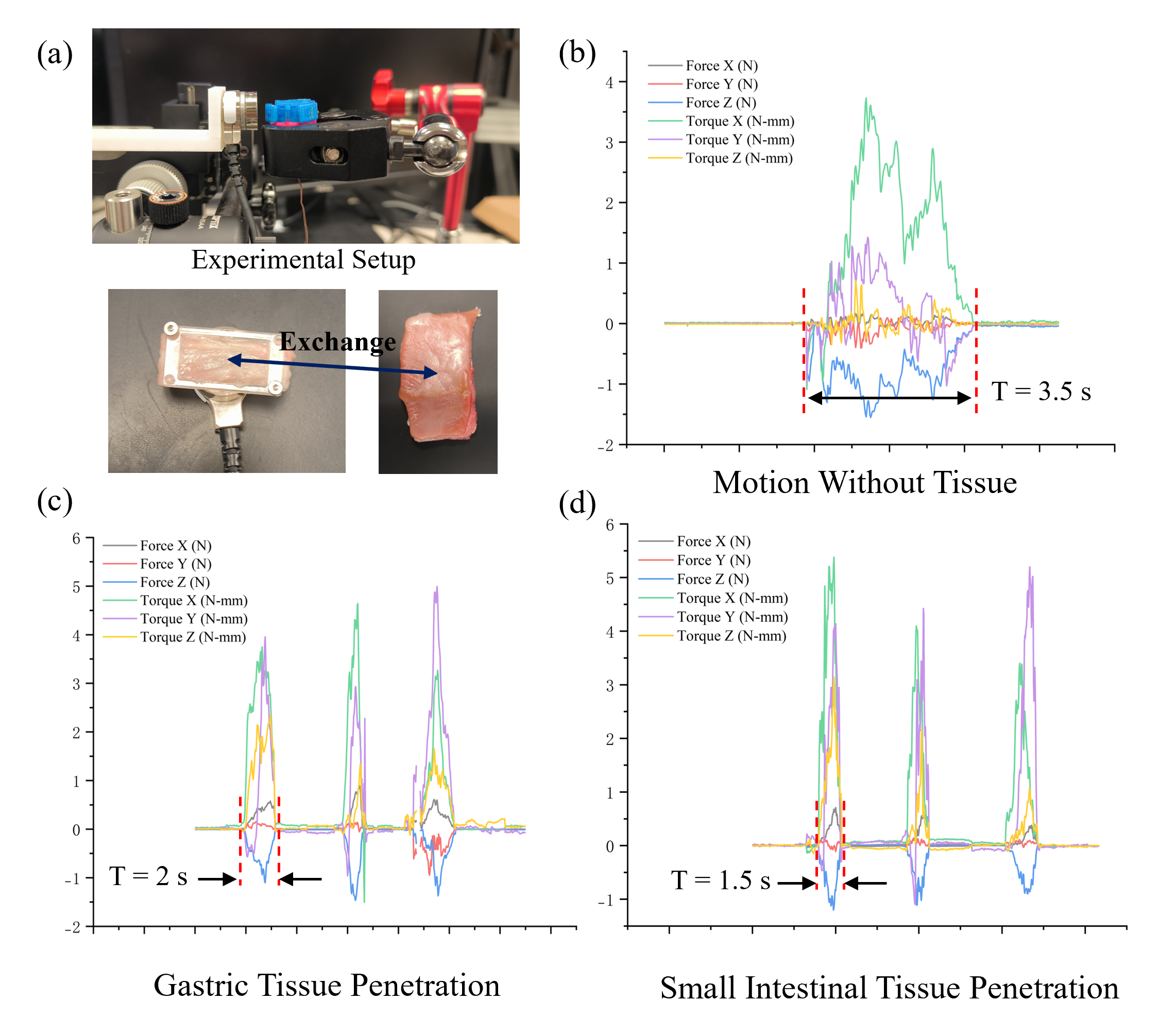}
  \caption{Force and penetration characterization of the kiri-capsule. (a) Experimental setup with ATI Nano17 force sensor and tissue fixation jig. (b) Force measurement during rotation without tissue contact ($T=3.5$~s), showing dominant $Z$-axis force. (c) Penetration forces measured in porcine gastric tissue ($T=2$~s), with forces in the $X$- and $Y$-axes rising to $0.5$--$2.0$~N. (d) Penetration forces in porcine small intestine ($T=1.5$~s), with lower thresholds of $0.3$--$1.0$~N.}
  \label{fig10}
\end{figure}

To characterize the interaction forces during deployment and penetration, calibration experiments were performed using a multi-axis force sensor. The experimental setup is shown in Fig.~\ref{fig10}a. An ATI Nano17 force/torque sensor (ATI Industrial Automation, NC, USA) was mounted on a multi-degree-of-freedom platform to ensure proper alignment between the capsule and the sensing surface. The fixture allowed adjustment to measure either axial or radial forces. For tissue-contact experiments, porcine gastric and small intestinal specimens were clamped to the sensor plane using a custom-designed holder. This setup enabled consistent measurement of contact and penetration forces during capsule actuation.

The baseline motion without tissue contact is shown in Fig.~\ref{fig10}b. During a 3.5~s rotation, the $Z$-axis force exhibited dominant fluctuations, while the $X$- and $Y$-axis forces remained relatively small. This confirmed that the motor-driven deployment and scraping primarily generated axial loads in the absence of tissue interaction.

When the capsule was actuated against gastric tissue (Fig.~\ref{fig10}c), the measured forces along the $X$- and $Y$-axes increased significantly, indicating penetration of the kirigami flaps into the tissue. The measured penetration forces fell within the expected physiological range for gastric biopsy (typically $0.5$--$2.0$~N), validating that the system achieves controlled engagement without excessive loading. For small intestinal tissue (Fig.~\ref{fig10}d), penetration occurred more rapidly, with force peaks observed at shorter actuation times ($\sim$1.5~s). The magnitude of the forces was generally lower, consistent with reported penetration thresholds of $0.3$--$1.0$~N for intestinal biopsy. These results confirm that the capsule can achieve effective insertion while maintaining safe force levels compatible with gastrointestinal tissues.

\section{Ex Vivo Biopsy Validation}

\begin{figure}[h]
  \centering
  \includegraphics[width=0.49\textwidth]{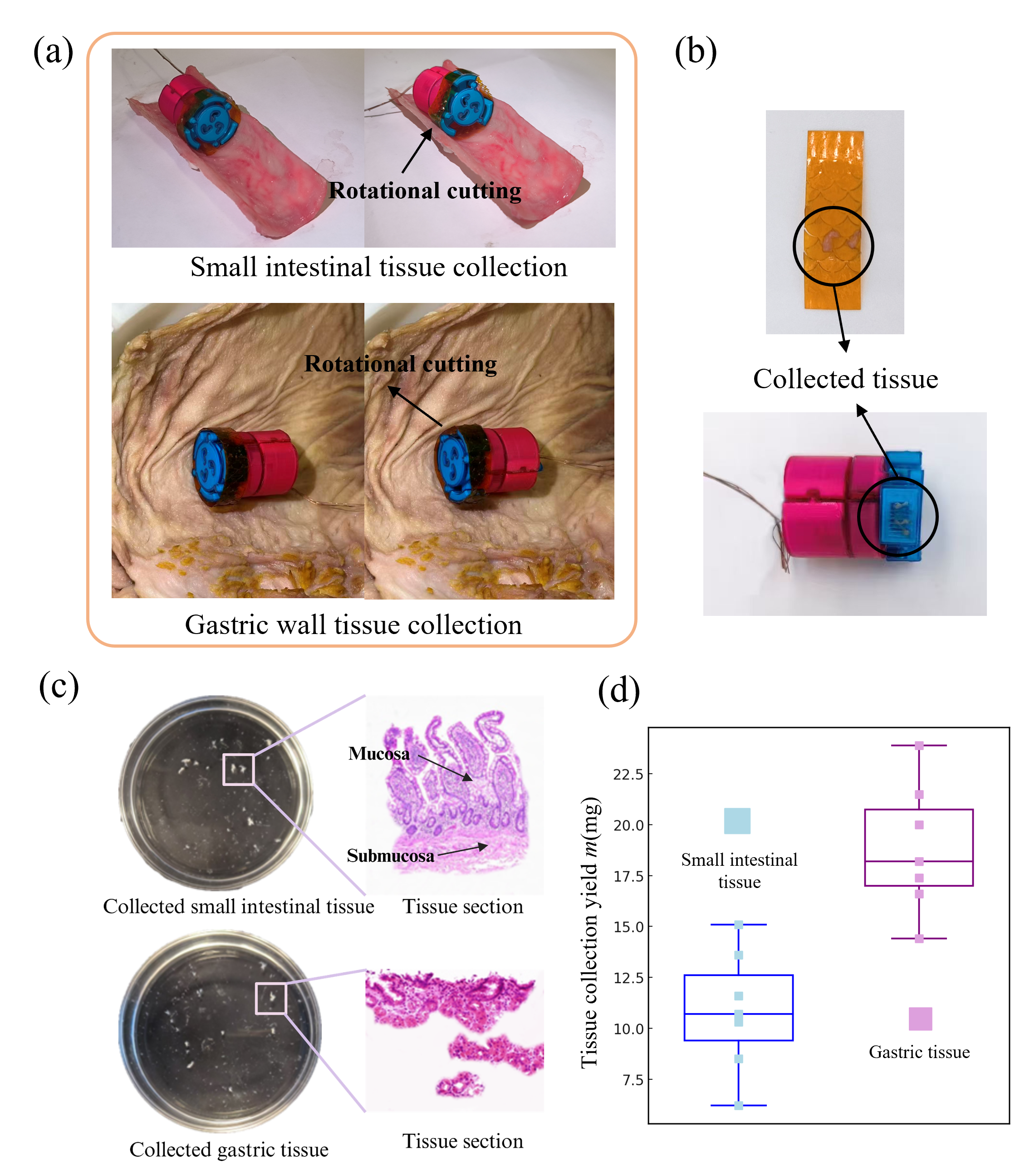}
  \caption{Ex vivo tissue collection validation. (a) Biopsy experiments in porcine small intestine (top) and gastric wall (bottom). (b) Collected tissue fragments attached to the kirigami capsule. (c) Retrieved tissue in Petri dishes with representative histological sections showing mucosa and submucosa. (d) Quantitative biopsy yield, with gastric tissue (median $=10.7$\,mg) and small intestinal tissue (median $=18.2$\,mg).}
  \label{fig11}
\end{figure}

To further assess the biopsy capability of the Kiri-Capsule, ex vivo experiments were conducted on porcine gastric wall and small intestinal tissues, which provide representative models of human GI morphology. Figure~\ref{fig11}a illustrates the experimental setup, where the capsule was actuated to deploy its kirigami flaps and perform controlled rotational scraping against the tissue surface. The upper panel depicts the sampling process in the small intestine. In contrast, the lower panel shows biopsy from the gastric wall, highlighting the adaptability of the system to different anatomical sites.  

Representative tissue samples collected during these procedures are shown in Fig.~\ref{fig11}b. Detached fragments adhered directly to the kirigami flaps and were subsequently transferred into the capsule’s internal cavities, demonstrating the system’s capacity for specimen capture and retention. Figure~\ref{fig11}c presents biopsy fragments retrieved into Petri dishes, alongside histological sections prepared using standard hematoxylin and eosin (H\&E) staining. These slices confirm that both mucosal and submucosal layers were successfully acquired, providing material suitable for diagnostic histopathology.  

The quantitative yield of collected specimens is summarized in Fig.~\ref{fig11}d. A total of seven biopsies were performed for each tissue type. Gastric biopsies produced smaller specimens (median $=10.7$\,mg, interquartile range (IQR) $=9.4$--$12.6$\,mg, range $=6.2$--$15.1$\,mg, mean $\approx 10.9$\,mg), consistent with the thicker and denser gastric wall. By contrast, small intestinal biopsies yielded significantly larger samples (median $=18.2$\,mg, IQR $=17.0$--$20.8$\,mg, range $=14.4$--$23.9$\,mg, mean $\approx 18.9$\,mg), reflecting the thinner and more compliant intestinal tissue. Data are presented as median~$\pm$~25--75\% confidence intervals (boxes), with whiskers showing the minimum and maximum values; cross markers indicate group means.  

These results confirm that the Kiri-Capsule can reproducibly acquire biopsy specimens of diagnostic relevance from both the stomach and the intestine. The controlled penetration depth, consistent tissue yield, and preserved histological integrity highlight the capability of the kirigami mechanism to achieve minimally invasive biopsy performance comparable to conventional forceps while maintaining the advantages of an ingestible capsule platform.


\section{Conclusion and Future Work}

In this work, we presented the design, fabrication, and validation of the Kiri-Capsule, a kirigami-inspired biopsy capsule robot for minimally invasive gastrointestinal diagnostics. By integrating a dual-cam expansion mechanism with a PI-based kirigami surface, the capsule achieves controlled deployment of sharp flaps and subsequent rotational scraping for tissue acquisition. Experimental validation confirmed the feasibility of this approach: tensile and angular characterization verified the mechanics of the kirigami sheets, penetration depth tests demonstrated shallow and safe insertion into soft tissue, and force calibration indicated actuation loads within clinically acceptable ranges. Ex vivo experiments on porcine stomach and small intestine tissues further validated the capsule’s ability to reliably collect biopsy samples of sufficient diagnostic quality.

Despite these promising results, several limitations remain. The current prototype requires a tethered power supply, restricting mobility and clinical applicability. Future work will focus on realizing untethered operation, for example, by adopting magnetic actuation to replace the stepper motor. Moreover, the present design supports only single-site sampling, and specimens may be mixed when multiple biopsies are attempted. To address this, future designs will explore multi-segment kirigami structures with independent actuators to enable spatially separated, multi-target biopsies. Additional improvements will also include optimizing capsule geometry, refining kirigami patterning, and investigating alternative biocompatible materials with enhanced resilience. These efforts will move the Kiri-Capsule closer to clinical translation as a safe, effective, and fully untethered biopsy platform for gastrointestinal diagnostics.

\addtolength{\textheight}{-12cm}   



\bibliographystyle{IEEEtran}


\end{document}